\documentclass[10pt,twocolumn,letterpaper]{article}

\usepackage{iccv}
\usepackage{times}
\usepackage{epsfig}
\usepackage{graphicx}
\usepackage{amsmath}
\usepackage{amssymb}

\usepackage{booktabs}
\usepackage{tabularx}
\newcolumntype{Y}{>{\centering\arraybackslash}X}
\usepackage{xspace}
\usepackage{enumitem}
\usepackage{microtype}
\usepackage{adjustbox}

\usepackage[pagebackref=true,breaklinks=true,letterpaper=true,colorlinks,bookmarks=false]{hyperref}

 \iccvfinalcopy % *** Uncomment this line for the final submission

\def\iccvPaperID{2353} % *** Enter the ICCV Paper ID here

\ificcvfinal\fi
\begin{document}

%%%%%%%%% TITLE

\title{Revisiting Point Cloud Classification: A New Benchmark Dataset and \\ Classification Model on Real-World Data}
\vspace{-0.2cm}
\author{Mikaela Angelina Uy$^1$~~~Quang-Hieu Pham$^2$~~~Binh-Son Hua$^3$~~~Duc Thanh Nguyen$^4$~~~Sai-Kit Yeung$^{1}$
\vspace{0.2cm}\\
$^1$Hong Kong University of Science and Technology~~~$^2$Singapore University of Technology and Design\\$^3$The University of Tokyo~~~$^4$Deakin University
\vspace{-0.2cm}\\
}

\maketitle
\begin{abstract}
\vspace{-0.3cm}
%Deep learning techniques for point cloud data has demonstrated great potentials in solving classical problems in 3D computer vision such as 3D object classification and segmentation. Several recent 3D object classification methods reported state-of-the-art performance on CAD model datasets such as ModelNet40 with high accuracy ($\sim$92\%). Despite such impressive results, in this paper, we argue that object classification is still a challenging task when objects are framed with real-world settings. To prove this, we create a new real-world point cloud object dataset based on popular indoor scene datasets including SceneNN and ScanNet. As objects from real-world scans are often cluttered with background and/or being partial due to occlusions, our dataset poses great challenges to existing point cloud classification techniques. We then conduct a comprehensive benchmark of existing point cloud object classification techniques on our new dataset. Based on the experimental results, we show that there remains three key open problems for point cloud object classification. Towards solving such challenging problems, we propose a new point cloud classification neural network that can address classifying objects with background. Our proposed method can outperform the state-of-the-art performance on our realistic object dataset.
Deep learning techniques for point cloud data have demonstrated great potentials in solving classical problems in 3D computer vision such as 3D object classification and segmentation. Several recent 3D object classification methods have reported state-of-the-art performance on CAD model datasets such as ModelNet40 with high accuracy ($\sim$92\%). Despite such impressive results, in this paper, we argue that object classification is still a challenging task when objects are framed with real-world settings. To prove this, we introduce ScanObjectNN, a new real-world point cloud object dataset based on scanned indoor scene data. From our comprehensive benchmark, we show that our 
dataset poses great challenges to existing point cloud classification techniques as objects from real-world scans are often cluttered with background and/or are partial due to occlusions. We identify three key open problems for point cloud object classification, and propose new point cloud classification neural networks that achieve state-of-the-art performance on classifying objects with cluttered background. Our dataset and code are publicly available in our project page \footnote{\url{https://hkust-vgd.github.io/scanobjectnn/}}.
\end{abstract}

\vspace{-0.5cm}
\section{Introduction}
\vspace{-0.1cm}
The task of understanding our real world has achieved a great leap in recent years. The rise of powerful computational resources such as GPUs and the availability of 3D data from depth sensors have accelerated the fast-growing field of 3D deep learning. Among various 3D data representations, point clouds are widely used in computer graphics and computer vision thanks to their simplicity. Recent works have shown great promises in solving classical scene understanding problems with point clouds such as 3D object classification and segmentation.

However, the current progress on classification with 3D point clouds has witnessed a trend of performance saturation. For example, many recent object classification methods have reported very high accuracies in 2018, and the trend of bringing the accuracy towards perfection is still ongoing. This phenomenon inspires us to raise a question on whether problems such as 3D object classification have been totally solved, and to think about how to move forward.

To answer this question, we perform a benchmark of existing point cloud object classification techniques with both synthetic and real-world data. 
For synthetic objects, we use ModelNet40~\cite{wu-3dshapenets-cvpr15}, the most popular dataset in point cloud object classification that contains about 10,000 CAD models.
To support the investigation of object classification methods on real-world data, we introduce ScanObjectNN, a new point cloud object dataset from the state-of-the-art scene mesh datasets SceneNN~\cite{hua-scenenn-3dv16} and ScanNet~\cite{dai2017scannet}. Based on the initial instance segmentation from the scene datasets, we manually filter and select objects for 15 common categories, and further enrich the dataset by considering additional object perturbations. 

Our study shows that while the accuracy with CAD data is reaching perfection, learning to classify a real-world object dataset is still a very challenging task. By analyzing the benchmark results, we identify three open issues that are worth to further explore for future researches. First, classification models trained on synthetic data often do not generalize well to real-world data such as point clouds reconstructed from RGB-D scans~\cite{hua-scenenn-3dv16,dai2017scannet}, and vice versa. Second, challenging in-context and partial observations of real-world objects are common due to occlusions and reconstruction errors; for example, they can be found in window-based object detectors~\cite{song-deep-cvpr16} in many robotics or autonomous vehicle applications. Finally, how to handle background effectively when they appear together with objects due to clutter in the real-world scenes. 

%in real-world settings, objects are often cluttered to background when they are observed in a scene. 

%Third, real-world data is imperfect; objects often appear partial due to occlusions and reconstruction errors. In fact, such challenging in-context and partial observations of objects are not uncommon; for example, they can be found in window-based object detectors~\cite{song-deep-cvpr16} in many robotics or autonomous vehicle applications. 

As our dataset opens up opportunities to tackle such open problems in real-world object classification, we also present a new method for point cloud object classification that can improve upon the state-of-the-art results on our dataset by jointly learning the classification and segmentation tasks in a single neural network. 

In summary, we make the following contributions:
\vspace{-0.2cm}
\begin{itemize}[leftmargin=*,itemsep=0ex]
    \item A new object dataset from meshes of scanned real-world scene for training and testing point cloud classification,
	\item A comprehensive benchmark of existing object classification techniques on synthetic and real-world point cloud data,
	\item A new network architecture that is able to classify objects observed in a real-world setting by a joint learning of classification and segmentation.
\end{itemize}

%The layout of our paper is as follows. We review related works in Section~\ref{sec:relatedworks}. Our dataset is presented in Section~\ref{sec:dataset}. We benchmark existing object classification techniques on our dataset in Section~\ref{sec:benchmarking}. Our proposed network to classify objects with background is presented in Section~\ref{sec:proposednetwork}. Section~\ref{sec:conclusion} concludes our paper and provides remarks. 
\section{Related Works}
\label{sec:relatedworks}
\vspace{-0.1cm}
% Dataset making can be made semi-automated with active learning~\cite{collins-dataset-active-eccv08,yu-lsun-ar15}. Instead of manually labeling all images, in this approach, a subset of images are first labelled and then used to train an initial classifier. The classifier is used for spliting the unlabeled images into a positive, negative, and unlabeled set. The most uncertain results are then presented to human for labeling, and the newly labeled images are incorporated into the training set. This process repeats until all images are labeled.

% TODO:
% Salient Objects in Clutter: Bringing Salient Object Detection to the
% Real-world objects are more memorable than photographs of objects
% Learning to Manipulate Unknown Objects in Clutter by Reinforcement
% Multi-view hypotheses transfer for enhanced object recognition in clutter
% Using Spin Images for Efficient Object Recognition in Cluttered 3D
% https://www.acin.tuwien.ac.at/en/vision-for-robotics/software-tools/object-clutter-indoor-dataset/
% Recognizing Objects in-the-Wild: Where do we Stand?

% Link to semi-supervised learning: train object classification to output object segmentation (background)

In this paper, we focus on object classification with point cloud data, which has advanced greatly in the past few years. We briefly discuss the related works and their datasets, below.
\vspace{-0.7cm}
\paragraph{Object Classification on Point Clouds.}
Early attempts to classifying point clouds were developed by adapting ideas from deep learning on images, \eg, using multiple view images~\cite{su15mvcnn, yu-multiview-cvpr18, yavar-spnet-accv18,kanezaki-rotationnet-cvpr18}, or applying convolutions on 3D voxel grids~\cite{maturana-voxnet-iros15, wu-3dshapenets-cvpr15}. While it seems natural to extend the convolution operations from 2D to 3D, it is shown that performing convolutions on a point cloud is not a trivial task~\cite{qi-pointnet-cvpr17,zaheer-deepsets-nips17}. The difficulty stems from the fact that a point cloud has no well-defined order of points on which convolutions can be performed. Qi et al.~\cite{qi-pointnet-cvpr17} addressed this problem by learning global features of point clouds using a symmetric function that is invariant to the order of points. Alternatively, some other methods proposed to learn local features from convolutions, \eg,~\cite{qi2017pointnetplusplus, li-pointcnn-ar18, hua-pointwise-cvpr18, wang2018edgeconv, xu2018spidercnn, hermosilla2018monte, ben20183dmfv,li-sonet-cvpr18,shen2018kcnet,domi-feature-wacv18} or from autoencoders~\cite{yang-foldingnet-cvpr18}. There are also methods jointly learning features from point clouds and multi-view projections~\cite{you-joint-mm18}. It is also possible to treat point clouds and views as sequences~\cite{liu2018point2seq, han-y2seq2seq-aaai19, han-3d2seqview-tip19}, or to use unsupervised learning~\cite{han-unsupervised-aaai19}. 

%In this work, we focus on learning techniques that only takes a point cloud as input for the object classification task.

Recent works demonstrate very competitive and compelling performances on standard datasets. For example, the gap between state-of-the-art methods such as SpecGCN \cite{wang2018local}, SpiderCNN \cite{xu2018spidercnn}, DGCNN \cite{wang2018edgeconv}, PointCNN \cite{li-pointcnn-ar18} is less than 1\% on ModelNet40 dataset~\cite{wu-3dshapenets-cvpr15}. In the online leaderboard maintained by the authors of ModelNet40, the accuracy of the object classification task is reaching perfection, with ~92\% for point cloud methods~\cite{li-pointcnn-ar18, wang2018edgeconv, xu2018spidercnn, liu2018point2seq}.

%This motivates us to ask the question whether point cloud learning should be considered solved, or which aspects has been missing in the current literature, which leads to our investigation of the current benchmark datasets. We focus on object classification, the classical scene understanding task that all works about point cloud deep learning reported their performance.
\vspace{-0.3cm}
\paragraph{Object Datasets.}
There are a limited number of datasets that can be used to train and test 3D object classification methods. ModelNet40 was originally developed by Wu et al.~\cite{wu-3dshapenets-cvpr15} for learning a convolutional deep-belief network to model 3D shapes represented in voxel grids. Objects in ModelNet40 are CAD models of 40 common categories such as airplane, motorbike, chair and table, to name a few. This dataset has been a common benchmark for point cloud object classification~\cite{qi-pointnet-cvpr17}. ShapeNet~\cite{chang-shapenet-2015} is an alternative large-scale dataset of 3D CAD shapes with approximately $51,000$ objects in 55 categories. However, this set is usually used for benchmarking part segmentation.

So far, object classification on ModelNet40 is done with the assumption that objects are clean, complete, and free from any background noise. Unfortunately, this assumption is not often held in practice. It is common to see incomplete (partial) objects due to the imperfection of 3D reconstruction. In addition, objects in real-world settings are often scanned when being placed in a scene, which makes them appear in a clutter, and thus may be attached with background elements. A potential treatment is to remove such background using human annotators~\cite{nguyen-anno-tvcg17}. However, this solution is tedious, prone to errors, and subjective to the experience of annotators. Other works synthesize challenges on CAD data by introducing noise simulated by Gaussians~\cite{Bobkov2018NoiseResistantDL, garcia2017} or created with a parametic model~\cite{Chandler2016MitigationOE} to mimic real world scenarios. Recently, the trend of sim2real~\cite{bewley2019sim2real} also aims to bridge the gap between synthetic and real data.

%Instead, we present a new dataset that mimics real world scenarios in object classification. Our goal is to make the neural networks for object classification more robust to imperfect point clouds. Our dataset is usable for both training and testing.

Prior to our work, there are also a few datasets of real-world object scans~\cite{mark-sydney-acra13,choi-dataset-ar16,calli-ycb-robotics17} but most are small in scale and are not suitable for training object classification networks, which often have thousands of parameters. For example, in robotics, Sydney urban objects dataset~\cite{mark-sydney-acra13} contains only 631 objects of 26 categories captured by a LiDAR camera, which is mainly used for evaluation~\cite{maturana-voxnet-iros15, ben20183dmfv} but not for training. Some datasets~\cite{singh-bigbird-icra14,calli-ycb-robotics17} are captured in controlled environment which might greatly differ from real-world scenes.
Choi et al.~\cite{choi-dataset-ar16} proposed a dataset of more than 10,000 object scans in the real world. However, not all of their scans can be successfully reconstructed; the online repository by the authors also provided only about 400 reconstructed objects.
RGB-D and 3D scene meshes datasets~\cite{hua-scenenn-3dv16,dai2017scannet,armeni_cvpr16,song-sunrgbd-cvpr15,nathan-nyu-eccv12} have more objects that are reconstructed along with the scenes, but such objects are often considered in a scene segmentation or object detection task, and not under an object classification setup.  
RGBD-to-CAD object classification challenge~\cite{hua-objectnn-shrec17, pham-shrec18} provides an object dataset that mixes CAD models and real-world scans. Its goal is to classify RGB-D objects such that a retrieval can be done to find similar CAD models. However, several categories are ambiguous, and objects are supposed to be well segmented before classification.
ScanNet~\cite{dai2017scannet} has a benchmark on 3D object classification with partially scanned objects. However, this dataset is designed for volume-based object classification~\cite{qi-volumetric-cvpr16}, and there are quite few techniques that report their results with this data.

%\cCH{To check if miss any dataset paper?}
%Here our goal differs in that we report the quantitative evaluations of various object classification techniques that take point cloud data as input. 
\begin{figure*}[t]
	\begin{center}
		\includegraphics[width=0.95\linewidth]{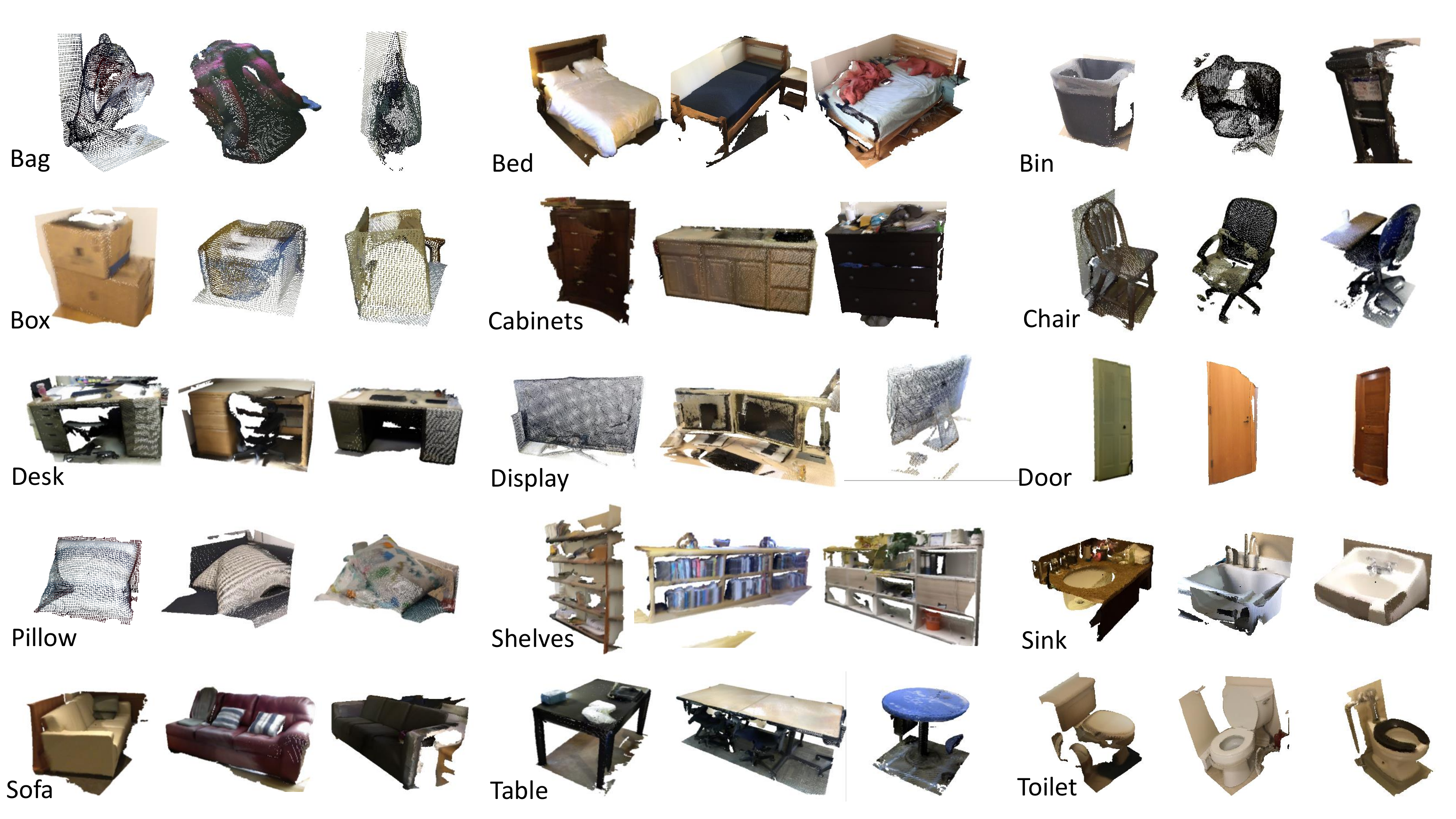}
	\end{center}
	\vspace{-0.25cm}
	\caption{Sample objects from our dataset.}
	\label{fig:object_scans}
	\vspace{-0.2cm}
\end{figure*}

%\vspace{-0.2cm}
\section{Benchmark Data}
\label{sec:dataset}
\vspace{-0.2cm}
%Deep neural networks on 3D geometric data have been progressing rapidly in recent years. Early attempts represented 3D data in regular grids (voxels)~\cite{wu-3dshapenets-cvpr15} or in multi-view images~\cite{qi-volumetric-cvpr16}. These representations enable straightforwardly applying convolutions. However, they require data quantization (\eg, voxelization) and may obscure important features from the original form of the data. In 2017, Qi \etal introduced PointNet~\cite{qi-pointnet-cvpr17}, the seminal work applying deep neural networks directly on 3D point clouds. This kick-started a fast growth of the field and showed great promise in the classical task of object classification. 

Our goal is to quantitatively analyze the performances of existing object classification methods on point clouds. We split our task into two parts: benchmarking with synthetic data and with real-world data. 

\subsection{Synthetic Data - ModelNet40}
\vspace{-0.1cm}
For synthetic data, we experiment with the well-known ModelNet40 dataset~\cite{wu-3dshapenets-cvpr15}. 
This set is a collection of CAD models with 40 object categories. The dataset includes 9,840 objects for training and 2,468 objects for testing. The objects in ModelNet40 are synthetic, and thus are complete, well-segmented, and noise-free. 
In this experiment, we use the uniformly dense point cloud variant as preprocessed by Qi et al.~\cite{qi-pointnet-cvpr17}. Each point cloud is randomly sampled to 1024 points as input to the networks unless otherwise stated. The point clouds are centered at zero, and we use local coordinates $(x, y, z)$ normalized to $[-1, 1]$ as point attributes. We follow the default train/test split, and use the default parameters as in the original implementations of the methods.
Our benchmark is performed with a NVIDIA Tesla P100 GPU.
We re-trained PointNet~\cite{qi-pointnet-cvpr17}, PointNet++~\cite{qi2017pointnetplusplus}, PointCNN~\cite{li-pointcnn-ar18}, Dynamic Graph CNN (DGCNN)~\cite{wang2018edgeconv}, 3D modified Fisher Vector (3DmFV)~\cite{ben20183dmfv}, and SpiderCNN~\cite{xu2018spidercnn}. For remaining methods, we provided the results reported in the original papers. We additionally report each method's best performance when provided with additional information such as point normals.
The results are shown in Table~\ref{tab:benchmark_modelnet}. It can be observed that the performance of recent methods is becoming incremental, and fluctuates around 92\%. 
This saturating score inspires us to revisit the object classification problem: Can classification methods trained on ModelNet40 perform well on real-world data? Or is there still room for more research problems to be explored?
% this almost perfect performance on CAD models generalized to real-world scenarios?
%This raises the question whether classification methods trained on ModelNet40 can perform well on real-world data.

\begin{table}
	\begin{center}
  		\small
		\begin{tabular}{c|c|c}
			\toprule
			Method& Avg. Class & Overall\\
			& Accuracy & Accuracy \\
			\midrule
			ECC \cite{Simonovsky2017ecc}& 83.2 & 87.4\\
			PointNet \cite{qi-pointnet-cvpr17}& 86.2 & 89.2\\
			DeepSets \cite{zaheer-deepsets-nips17}&  - & 90.0\\
			Flex-Convolution \cite{accv2018/Groh}&-&90.2\\
			Kd-Net \cite{KlokovL17}& 88.5 & 90.6 (91.8 *)\\						
			PointNet++ \cite{qi2017pointnetplusplus}& 87.8 & 90.7 (91.9 w/ normal)\\
			SO-Net \cite{li-sonet-cvpr18} & 87.3 & 90.9 (93.4 w/ normal) \\
			KCNet \cite{shen2018kcnet} & - & 91\\
			3DmFV \cite{ben20183dmfv} & 86.3 & 91.4\\
			SpecGCN \cite{wang2018local}& - & 91.5 (92.1 w/ normal)\\
			SpiderCNN \cite{xu2018spidercnn} & 86.8 & 90.0 (92.4 w/ normal)\\
			DGCNN \cite{wang2018edgeconv} & 90.2 & 92.2 \\
			PointCNN \cite{li-pointcnn-ar18} & 88.8 & 92.5 \\
			\bottomrule			
		\end{tabular}
	\end{center}
	\vspace{-0.1cm}
	\caption{Baseline results on ModelNet40 dataset for point cloud classification. Inputs are point coordinates, unless otherwise stated; * denotes the use of more input points (32K).}
	\vspace{-0.5cm}
	\label{tab:benchmark_modelnet}
\end{table}
%\vspace{-0.4cm}
%\begin{table}
%	\begin{center}
%		\small
%		\begin{tabular}{c|c|c}
%			\toprule
%			Method& Avg. Class & Overall Accuracy\\
%			& Accuracy & \\
%			\midrule
%			ECC \cite{Simonovsky2017ecc}& 83.2 & 87.4\\
%			PointNet \cite{qi-pointnet-cvpr17}& 86.2 & 89.2\\
%			DeepSets \cite{zaheer-deepsets-nips17}&  - & 90.0\\
%			Flex-Convolution \cite{accv2018/Groh}&-&90.2\\
%			Kd-Net \cite{KlokovL17}& 88.5 & 90.6 \\						
%			PointNet++ \cite{qi2017pointnetplusplus}& 87.8 & 90.7\\
%			SO-Net \cite{li-sonet-cvpr18} & 87.3 & 90.9 \\
%			KCNet \cite{shen2018kcnet} & - & 91\\
%			3DmFV \cite{ben20183dmfv} & 86.3 & 91.4\\
%			SpecGCN \cite{wang2018local}& - & 91.5\\
%			SpiderCNN \cite{xu2018spidercnn} & 86.8 & 90.0\\
%			DGCNN \cite{wang2018edgeconv} & 90.2 & 92.2 \\
%			PointCNN \cite{li-pointcnn-ar18} & 88.8 & 92.5 \\
%			\bottomrule			
%		\end{tabular}
%	\end{center}
%	\caption{Baseline results on ModelNet40 dataset for point cloud classification}
%	\label{tab:benchmark_modelnet}
%\end{table}

\begin{table*}
	\begin{center}
  		\small
		\begin{tabular}{c|c|c|c|c|c|c|c|c|c|c|c|c|c|c|c}
			\toprule
			\textbf{Class} & Bag & Bed & Bin & Box & Cabinet & Chair & Desk & Display & Door & Pillow & Shelf & Sink & Sofa & Table & Toilet \\
			\midrule
			%\hline
			\textbf{\#Objects} & 78 & 135 & 201 & 127 & 347 & 395 & 149 & 181 & 221 & 105 & 267 & 118 & 254 & 242 & 82 \\
			\bottomrule			
		\end{tabular}
	\end{center}
%	\vspace{-0.1cm}
	\caption{Classes and objects in our dataset.}
	\label{tab:dataset}
%	\vspace{-0.1cm}
\end{table*}

%\begin{figure}[t]
%    \begin{center}
%    \includegraphics[width=1\linewidth]{figures/real_vs_cad.pdf}
%    \end{center}
%	\caption{Example real-world scans and CAD models of some common object classes.}
%	\label{fig:CADvsReal}
%\end{figure}

\subsection{Real-World Data - ScanObjectNN}
\vspace{-0.1cm}
%However, it is not trivial to apply existing point cloud classification methods to real-world data. 
%Despite the impressive performance on CAD models, we suggest that on real-world objects will need further considerations. 
%As shown in Figure~\ref{fig:CADvsReal}, 
Objects obtained from real-world 3D scans are significantly different from CAD models due to the presence of background noise and the non-uniform density due to holes from incomplete scans/reconstructions and occlusions. This situation is often seen in sliding window-based object detection \cite{song-deep-cvpr16} in which a window may enclose an object of interest partially and also include background elements within the window. %This raises the question whether classification methods trained on ModelNet40 can perform well on real-world data.
%This is not trivial to apply existing point cloud classification methods to real-world data. 
Due to these properties, applying existing point cloud classification methods to real-world data may not produce the same good results as CAD models. 

\vspace{-0.1cm}
\subsubsection{Data Collection}
\vspace{-0.1cm}
To study this potential issue, we build a real-world object dataset based on two popular scene meshes datasets: SceneNN~\cite{hua-scenenn-3dv16} and ScanNet~\cite{dai2017scannet}. SceneNN has 100 annotated scenes with highly cluttered objects while ScanNet has a larger collection of 1513 indoor scenes.
From a total of more than 1600 scenes from SceneNN and ScanNet, we selected 700 unique scenes. We then manually examined each object, fixed inconsistent labels, and discard objects that are ambiguous, have low reconstruction quality, have unknown labels, are too sparse, and have too few instances to form a category for training. 
During categorization, we also took into account inter-class balancing to avoid any bias potentially coming from classes with more samples.

The results are $2902$ objects that are categorized into 15 categories.
The raw objects are represented by a list of points with global and local coordinates, normals, colors attributes and semantic labelsOther works synthesize challenges on CAD data by introducing noise simulated by Gaussians~\cite{Bobkov2018NoiseResistantDL, garcia2017} or created with a parametic model~\cite{Chandler2016MitigationOE}. Recently, the trend of sim2real~\cite{bewley2019sim2real} also aims to bridge the gap between synthetic and real data. 
As in the experiment with synthetic data, we sample all raw objects to 1024 points as input to the networks and all methods were trained using only the local $(x,y,z)$ coordinates. We will make our dataset publicly available for future research.
Table~\ref{tab:dataset} summarizes classes and objects in our dataset.
\vspace{-0.2cm}
\subsubsection{Data Enrichment}
\begin{figure}[t]
	\def\sc{0.49}
	\centering
	\begin{minipage}{\sc\linewidth}
		\includegraphics[width=\linewidth]{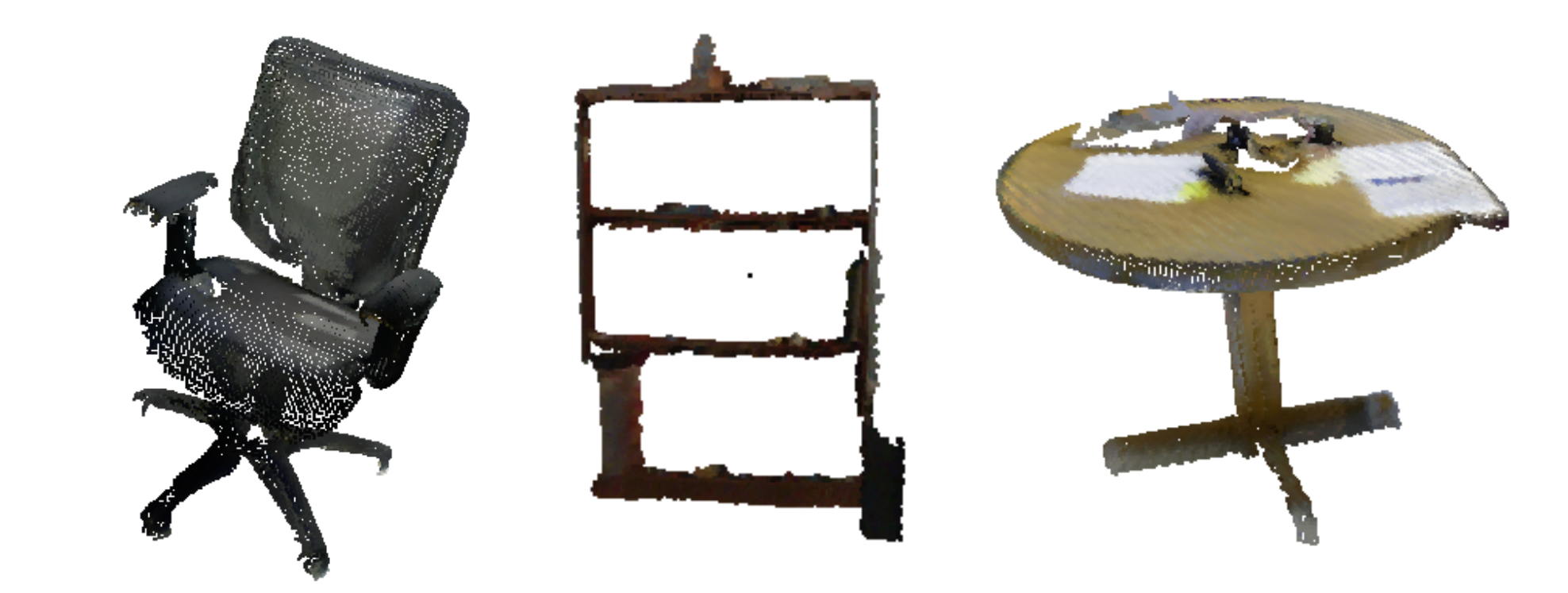}
		\centerline{\small(a) Objects only.}
	\end{minipage}
	\begin{minipage}{\sc\linewidth}
		\begin{center}
			\includegraphics[width=\linewidth]{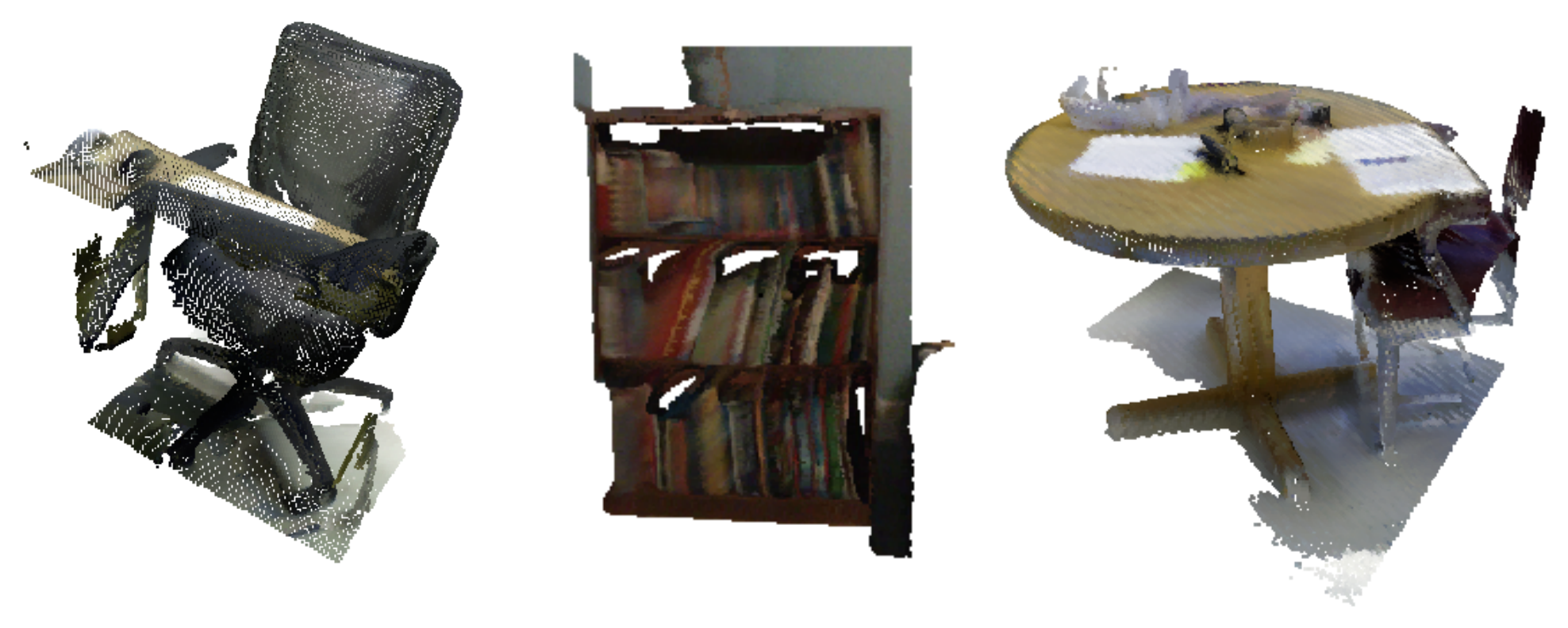}
			\centerline{\small(b) Objects with background.}
		\end{center}
	\end{minipage}
	\vspace{0.1in}
	\caption{Example objects from our dataset.}
	\vspace{-0.5cm}
	\label{fig:objects}
\end{figure}

\vspace{-0.1cm}
Based on the selected objects, we construct several variants that represent different levels of difficulty of our dataset. This allows us to explore the robustness of existing classification methods in more extreme real-world scenarios.
\vspace{-0.4cm}
\noindent \paragraph{Vanilla.} The first variant is referred to as \textbf{OBJ\_ONLY} which includes only ground truth segmented objects extracted from the scene meshes datasets. This variant has the closest form analogous to its CAD counterpart, and is used to investigate the robustness of classification methods to noisy objects with deformed geometric shape and non-uniform surface density. Sample objects of this variant are shown in Figure~\ref{fig:objects}(a).
\vspace{-0.4cm}
\noindent \paragraph{Background.} The previous variant assumes that an object can be accurately segmented before being classified. However, in real-world scans, objects are often presented in under-segmentation situations, i.e., background elements or parts of nearby objects are included, and accurate annotations for such under-segmentations are also not always available. Those background elements may provide the context where objects belong to, and thus would become a good hint for object classification, \eg, laptops often sit on desks. However, they may also introduce distractions which corrupt the classification, \eg, a pen may be under-segmented with a table where it sits on and thus could be considered as a part of the table rather than a separate object. To study these factors, we introduce a variant of our dataset where objects are attached with background data (\textbf{OBJ\_BG}). We determine such background by using the ground truth axis-aligned object bounding boxes. Specifically, given a bounding box, all points in the box are extracted to form an object. Sample objects with background are shown in Figure~\ref{fig:objects}(b).

%\vspace{-0.58cm}
\noindent \paragraph{Perturbed.} The given bounding boxes from the ground-truth tightly enclose the objects. However, in real-world scenarios bounding boxes may over- or under-cover, or even split objects. For example, in object detection techniques such as R-CNN~\cite{girshick-rcnn-cvpr14}, object category has to be predicted from a rough bounding box that localizes a candidate object. To simulate this challenge, we extend our dataset by translating, rotating (about the gravity axis), and scaling the ground truth bounding boxes before extracting the geometry in the box. We name the variants of these perturbations with a common prefix \textbf{PB}.

\begin{figure}[t]
	\centering
	\includegraphics[width=\linewidth]{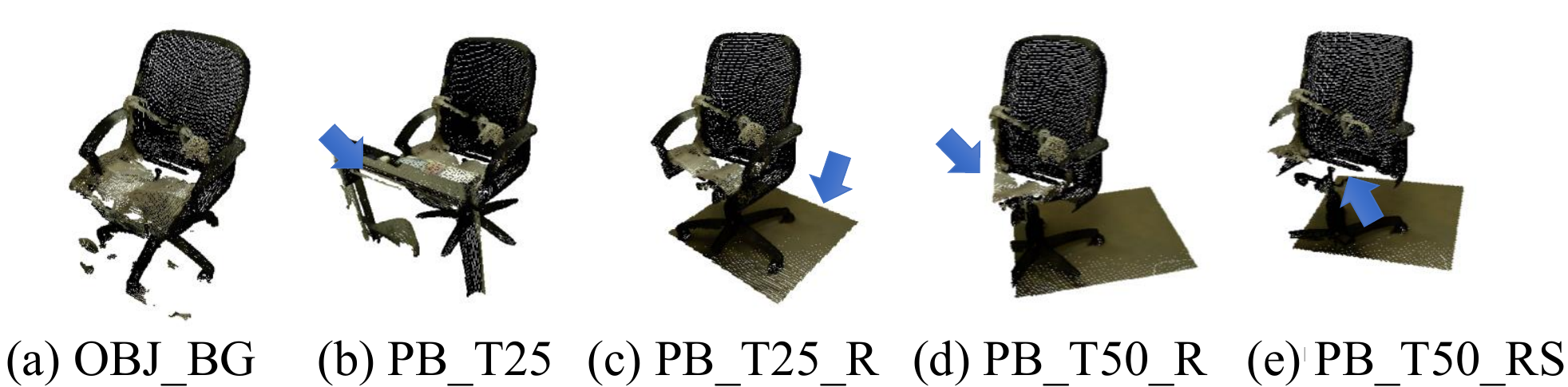}
	\caption{An object in different perturbation variants.}
	\label{fig:perturbations}
	\vspace{-0.2cm}
\end{figure}

The perturbations introduce various degrees of background and partiality to objects. In this work, we use four perturbation variants in the increasing order of difficulty: \textbf{PB\_T25}, \textbf{PB\_T25\_R}, \textbf{PB\_T50\_R}, and \textbf{PB\_T50\_RS}. 
Suffix \textbf{\_T25} and \textbf{\_T50} denote translation that randomly shifts the bounding box up to 25\% and 50\% of its size from the box centroid along each world axis. Suffix \textbf{\_R} and \textbf{\_S} denotes rotation and scaling. 
Each perturbation variant contains five random samples for each original object, resulting in up to $14,510$ perturbed objects in total. Since perturbation might introduce invalid objects, e.g., objects that are almost completely out of the bounding box of interest, we perform an additional check after perturbation by ensuring that at least 50\% of the original object points remain in the bounding box. Objects that do not satisfy this condition are discarded. Sample point clouds of these variants are shown in Figure~\ref{fig:perturbations}. 
More details about perturbing objects can be found in our supplementary material.

%We further scale our dataset as follows. Here we draw our inspiration from the imperfection of object classification in practice. For example, in a two-stage object detection technique~\cite{rcnn}, object label has to be predicted from a possibly shifted bounding box that localizes a candidate object. We simulate this behavior by perturbing the ground truth bounding boxes before extracting the geometry. The perturbation we performed includes translation, rotation about the gravity axis, and scale. Here we limit ourselves to rotation about the gravity axis since so far most of the representative techniques for point cloud classification are not designed to be rotation invariant. In other words, we assume that the floor plane can be reliably determined before an object is classified.

\vspace{-0.5cm}
\section{Benchmark on ScanObjectNN}
\label{sec:benchmarking}
\vspace{-0.1cm}
%Note that all point clouds are normalized to a unit ball and centered at the origin.

For a clearer picture of the maturity of point cloud-based object classification, we benchmark several representative methods on our dataset. We aim to identify the limitations of current works on real-world data. We choose 3DmFV \cite{ben20183dmfv}, PointNet \cite{qi-pointnet-cvpr17}, SpiderCNN \cite{xu2018spidercnn}, PointNet++ \cite{qi2017pointnetplusplus}, DGCNN \cite{wang2018edgeconv} and PointCNN\cite{li-pointcnn-ar18} as our representative works.

%Note that ModelNet40 contains only clean objects and are never introduced the concept of background.
%We then re-centered the remaining point cloud to the origin.
\vspace{-0.1cm}
\subsection{Training on ModelNet40}
\vspace{-0.1cm}
We first study the case when training is done on ModelNet40 and testing is done on ScanObjectNN. Since objects in ModelNet40 are standalone with no background objects, we also removed background in all our variants for fair evaluations. Furthermore, we only evaluated the current methods on 11 (out of 15) common classes between ModelNet40 and our dataset.
%To investigate the challenges raised by real-world objects themselves, we removed the background in perturbed objects. To make evaluations fair, we only evaluated the current methods on 11 (out of 15) common classes between ModelNet40 and our dataset. 
Please refer to the supplementary material for the details on these common classes.

Evaluation results are reported in Table~\ref{tab:modelnet_to_scan}. These results show that the current techniques trained on CAD models are not able to generalize to real-world data; all techniques achieved less than 50\% of accuracy. This is expected and is because of the fact that real-world objects and CAD objects are significantly different in their geometry. Real-world objects are often incomplete and partial due to construction errors and occlusions; their surfaces have low-frequency noise; object boundaries are inaccurate. These are in contrast to CAD objects, which are often clean and noise-free. We also found that the harder the data is (\ie more noise and partiality), the lower the performance is, and this is consistent for all techniques. In other words, knowledge learned from synthetic objects in ModelNet40 is not well transferable and/or applicable to real-world data.

\newcolumntype{R}[2]{%
	>{\adjustbox{angle=#1,lap=\width-(#2)}\bgroup}%
	l%
	<{\egroup}%
}
\newcommand*\rot{\multicolumn{1}{R{45}{1em}}}
\begin{table}
	\begin{center}
		\small
		\begin{tabular}{c|c|c|c|c|c}
			\toprule
			& \rot{OBJ\_ONLY} & \rot{PB\_T25} & \rot{PB\_T25\_R} & \rot{PB\_T50\_R} & \rot{PB\_T50\_RS}\\
			\midrule
			3DmFV \cite{ben20183dmfv}& 30.9 & 28.4 & 27.2 & 24.5 & 24.9 \\
			PointNet \cite{qi-pointnet-cvpr17}& 42.3 & 37.6 & 35.3 & 32.1 & 31.1 \\
			SpiderCNN \cite{xu2018spidercnn}& 44.2 & 37.7 & 34.5 & 31.7 & 30.9 \\					
			PointNet++ \cite{qi2017pointnetplusplus}& 43.6 & 37.8 & 37.2 & 33.3 & 32.0\\
			DGCNN \cite{wang2018edgeconv}& 49.3 & 42.4 & 40.3 & 36.6 & 36.8 \\
			PointCNN \cite{li-pointcnn-ar18}& 32.2 & 28.7 & 28.1 & 26.4 & 24.6 \\
			\bottomrule			
		\end{tabular}
	\end{center}
	\vspace{-0.2cm}
	\caption{Overall accuracy in \% on our dataset when training was done on ModelNet40. Note that for a fair comparison, background has been removed in all variants. The results show that training on CAD models and testing on real-world data is challenging. Most methods do not generalize well in this test.}
	\vspace{-0.25cm}
	\label{tab:modelnet_to_scan}
\end{table}

\begin{table}
	\begin{center}
		\small
		\begin{tabular}{c|c|c|c|c|c|c}
			\toprule
			& \rot{OBJ\_ONLY} & \rot{OBJ\_BG} & \rot{PB\_T25} & \rot{PB\_T25\_R} & \rot{PB\_T50\_R} & \rot{PB\_T50\_RS}\\
			\midrule
			3DmFV \cite{ben20183dmfv} & 73.8 & 68.2 & 67.1 & 67.4 & 63.5 & 63.0 \\
			PointNet \cite{qi-pointnet-cvpr17} & 79.2 & 73.3 & 73.5 & 72.7 & 68.2 & 68.2  \\
			SpiderCNN \cite{xu2018spidercnn} & 79.5 & 77.1 & 78.1 & 77.7 & 73.8 & 73.7 \\					
			PointNet++ \cite{qi2017pointnetplusplus} & 84.3 & 82.3 & 82.7 & 81.4  & 79.1 & 77.9 \\
			DGCNN \cite{wang2018edgeconv} & 86.2 & 82.8 & 83.3 & 81.5 & 80.0 & 78.1 \\
			PointCNN \cite{li-pointcnn-ar18} & 85.5 & 86.1 & 83.6 & 82.5 & 78.5& 78.5\\
			\bottomrule			
		\end{tabular}
	\end{center}
	\vspace{-0.2cm}
	\caption{Overall accuracy in \% when training and testing were done on ScanObjectNN. The training and testing are done on the same variant. With real-world data, the more background and partiality are introduced, the more challenging the classification task is.}
	\vspace{-0.4cm}
	\label{tab:ours}
\end{table}
\vspace{-0.1cm}
\subsection{Training on ScanObjectNN}
\vspace{-0.1cm}
In this experiment, we train and test the techniques on ScanObjectNN to demonstrate training on datasets with real-world properties should improve the performance in classifying real-world objects. We also analyze how different perturbations can affect the classification performance. We randomly split our dataset into two subsets: training (80\%) and test (20\%) set. We ensure that the training and test sets contain objects from different scenes so that similar objects do not occur in the same set, \eg same types of chairs can be found in the same room. We report the performance of all the techniques on the hardest split in Table~\ref{tab:ours}. Full performances on all splits are provided in our supplementary material.

For fair comparisons, we kept the same data augmentation process in all the methods (\eg, random rotation and per-point jitter). We trained the methods to convergence rather than selecting the best performance on the test set.\\ 

%\noindent\textbf{Baseline-Variant.} The 2nd column in Table~\ref{tab:ours} shows the overall performance of existing methods when trained on the simplest variant of our dataset (OBJ\_ONLY). The results in Table~\ref{tab:ours} clearly show that the classification accuracy on our dataset is substantially lower than that on ModelNet40 (in Table~\ref{tab:benchmark_modelnet}). This means that the problem of point cloud classification on real-world data is still open. In the following, we investigate different types of perturbations in the variants of our dataset.\\
\vspace{-0.2cm}
\noindent\textbf{Vanilla.} The 2nd column in Table~\ref{tab:ours} shows the overall performance of existing methods when trained on the simplest variant of our dataset (OBJ\_ONLY). This clearly shows that the classification accuracy increased significantly when training and testing are both done using ScanObjectNN versus when training is done using ModelNet40 (Table~\ref{tab:modelnet_to_scan} Column 2). However, we also notice an observable performance drop comparing to the pure synthetic setting in Table~\ref{tab:benchmark_modelnet}. This gives an important message: point cloud classification on real-world data is still open, a dataset with real-world properties can help, but further research is necessary to regain the high performance as in synthetic setting. 
%substantially lower than that on ModelNet40 (in Table~\ref{tab:benchmark_modelnet}). This means that the problem of point cloud classification on real-world data is still open. 
In the following, we investigate the performance change in different types of perturbations in our dataset.\\

\vspace{-0.2cm}
\noindent\textbf{Background.} As shown in Table~\ref{tab:ours} Columns 3-7, background makes strong impact to the classification performance of all methods. Specifically, except PointCNN \cite{li-pointcnn-ar18}, all methods performed worse on OBJ\_BG compared with OBJ\_ONLY. It can be explained by the fact that background elements could distract the learning in existing methods by confusing between foreground and background points. 
%PointCNN performed similarly on both OBJ\_ONLY and OBJ\_BG probably because of the use of X-Conv which diminishes the weights of background points. 
To further confirm the negative effect of having background objects, we conduct a control experiment using the hardest perturbation variant, \ie, PB\_T50\_RS. Table~\ref{tab:background} shows the overall accuracy of all existing models decrease when trained and tested \emph{with} the presence of background.\\

\begin{table}
	\begin{center}
  		\small
		\begin{tabular}{c|c|c|c|c}
			\toprule
			& \multicolumn{2}{c}{Ours} & \multicolumn{2}{c}{ModelNet40} \\
			& {w/o BG} & {w/ BG} & {w/o BG} & {w/ BG}\\
			\midrule
			3DmFV \cite{ben20183dmfv}& 69.8 & 63.0 & 54.1 & 51.5 \\
			PointNet \cite{qi-pointnet-cvpr17}& 74.4 & 68.2 & 60.4 & 50.9 \\
			SpiderCNN \cite{xu2018spidercnn}& 76.9 & 73.7 & 52.7 &  46.6 \\	
			PointNet++ \cite{qi2017pointnetplusplus}& 80.2 & 77.9 & 55.0 & 47.4 \\
			DGCNN \cite{wang2018edgeconv}& 81.5 & 78.1 & 58.7 & 54.7 \\
			PointCNN \cite{li-pointcnn-ar18}& 80.8 & 78.5 & 38.1 & 49.2 \\
			\bottomrule			
		\end{tabular}
	\end{center}
	\vspace{-0.2cm}
	\caption{Overall accuracy in \% when training on our hardest variant PB\_T50\_RS, with and without background (BG) points. Testing is done on the same variant of our dataset, and on ModelNet40. The second header indicates the results corresponding to the training set. The results show that (1) background impacts negatively to the classification performance, and (2) training on our real-world objects generalizes to CAD evaluation better than the opposite case.}
	\label{tab:background}
	\vspace{-0.4cm}
\end{table}

\begin{figure}[t]
	\begin{center}
		\includegraphics[width=0.85\linewidth]{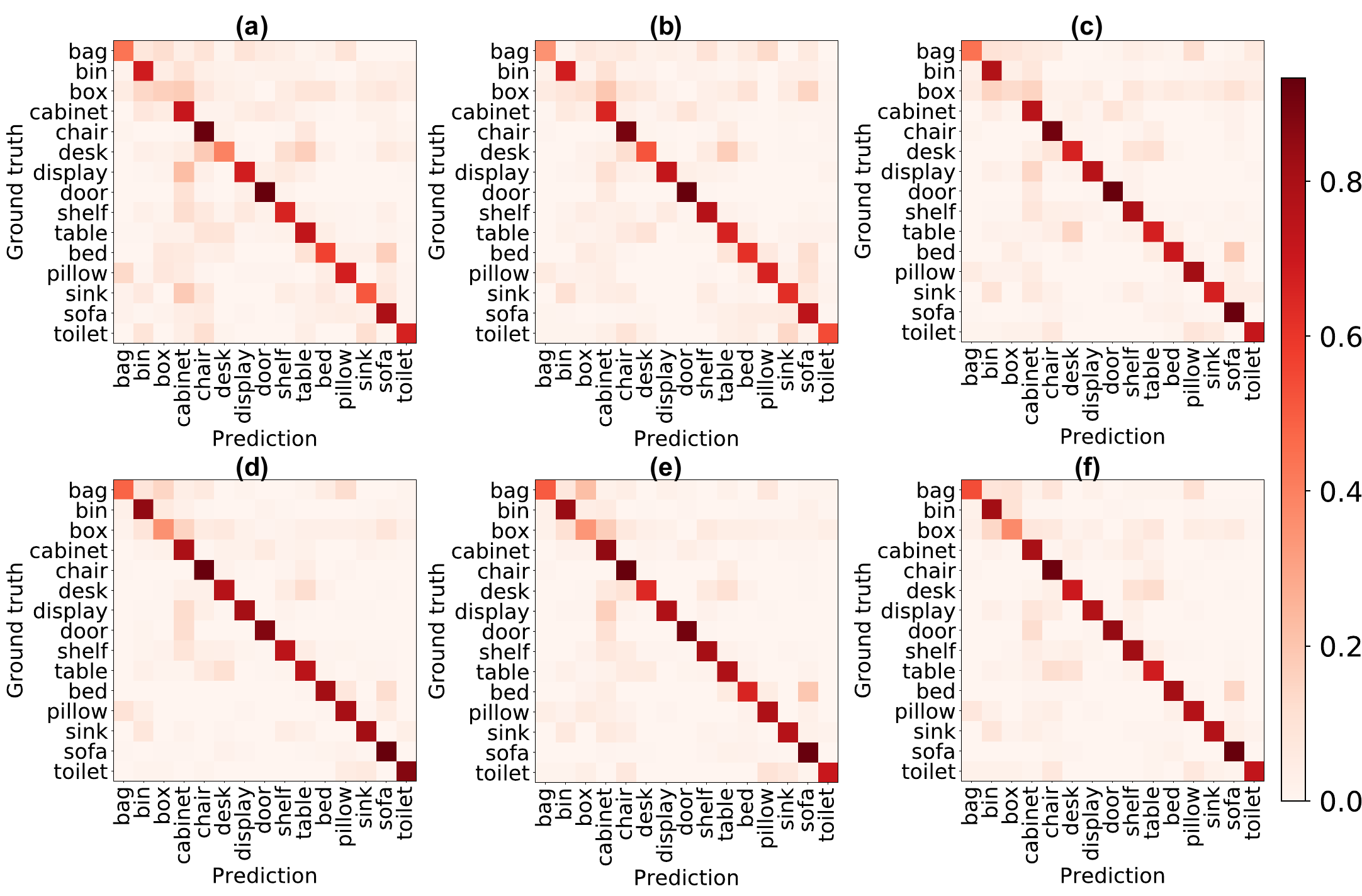}
	\end{center}
	\vspace{-0.4cm}
	\caption{Confusion matrices of (a) 3DmFV~\cite{ben20183dmfv}, (b) PointNet~\cite{qi-pointnet-cvpr17}, (c) SpiderCNN~\cite{xu2018spidercnn}, (d) PointNet++~\cite{qi2017pointnetplusplus}, (e) DGCNN~\cite{wang2018edgeconv} and (f) PointCNN~\cite{li-pointcnn-ar18} on our hardest PB\_T50\_RS. This shows that there are no major ambiguity issues among object classes in our dataset.}
	\label{fig:confusion_matrix_pbt50rs}
	\vspace{-0.4cm}
\end{figure}

\vspace{-0.2cm}
\noindent\textbf{Perturbation.}
Table~\ref{tab:ours} also shows the impact of perturbations to the classification performance (compared with Column 2). In this result, we observe that translation and rotation both make the classification performance decrease significantly, especially with larger perturbations that introduce more background and partiality. Scale further degrades the performance by a small gap. 
Figure~\ref{fig:confusion_matrix_pbt50rs} illustrates the confusion matrices of all methods on our hardest variant PB\_T50\_RS. It can be seen that there are no major ambiguity issues in our categories, and our dataset is challenging due to the high variations in real-world data.\\ 

%\noindent\textbf{Translation Perturbation.} Table~\ref{tab:ours} Columns 4-7 show the impact of translation to the classification performance (compared with Column 2). Translation produces partial objects. Moreover, with the presence of background, translation also deviates objects from the origin, which adds another difficulty in object alignment used in classification methods.\\

%\noindent\textbf{Rotation Perturbation.} Table~\ref{tab:ours} Column 5 shows that rotation perturbation decreases the classification performance (compared with Column 4). This is probably because rotating bounding boxes would add more background points and hence make more distractions to the learning process.\\

%\noindent\textbf{Scale Perturbation.} Table~\ref{tab:ours}-Column 7 shows that scaling also affects the classification task, specially when combined with rotation and translation. However, scale perturbation does not make significant impact on classification performance as so do rotation and translation ones. Scaling may include more background but this effect has also been considered in translation perturbation.\\

\vspace{-0.2cm}
\noindent\textbf{Generalization to CAD Data.} 
While it is shown that networks trained on synthetic data generalizes poorly to our dataset (Table~\ref{tab:modelnet_to_scan}), the reverse is not true. Here we tested the generalization capability of existing methods when trained on ScanObjectNN. In this experiment, all methods were trained on our PB\_T50\_RS (with and without background) and tested on ModelNet40. The results in the last two columns in Table~\ref{tab:background} clearly show that existing methods could generalize better when they were trained on real-world data (compared with the results in Table~\ref{tab:modelnet_to_scan}). Performance on individual classes are presented in Table~\ref{tab:per_class_avg_modelnet}. As shown in Table~\ref{tab:per_class_avg_modelnet}, lower accuracies are achieved on classes such as bed, cabinet, and desk, where complete structures are never observed in real scans because these objects are often situated adjacent to walls or near corners of rooms.
Therefore, we advocate using real-world data in training object classification because the generalization is shown to be much better.

\begin{table*}
	\begin{center}
  		\small
		\begin{tabular}{c|c|c|c|c|c|c|c|c|c|c|c}
			\toprule
			& cabinet & chair & desk & display & door & shelf & table & bed & sink & sofa & toilet\\
			\midrule
			3DmFV \cite{ben20183dmfv}&20.8&67.1&\textbf{8.1}&75.0&75.0&86.0&97.0&10.0&50.0&21.0&64.0\\
			PointNet \cite{qi-pointnet-cvpr17}&\textbf{2.8}&72.1&43.0&83.0&100.0&98.0&93.0&\textbf{4.0}&35.0&23.0&26.0\\
			SpiderCNN \cite{xu2018spidercnn}& 17.9 & 54.3 & 17.4 & 86.0 & 90.0 & 90.0 & 88.0 & \textbf{7.0} & 40.0 & 32.0 & 14.0\\
			PointNet++ \cite{qi2017pointnetplusplus}&18.9&71.4&12.8&94.0&45.0&79.0&88.0&\textbf{2.0}&45.0&14.0&35.0\\
			DGCNN \cite{wang2018edgeconv}&47.2&75.7&11.6&94.0&85.0&83.0&100.0&\textbf{9.0}&45.0&42.0&12.0\\
			PointCNN \cite{li-pointcnn-ar18}&42.5&77.9&24.4&76.0&20.0&92.0&76.0&\textbf{4.0}&35.0&24.0&19.0\\			
			\bottomrule			
		\end{tabular}
	\end{center}
	\vspace{-0.2cm}
	\caption{Per class average accuracy in \% on ModelNet40 when training was done on our PB\_T50\_RS. Low accuracies are highlighted.}
	\vspace{-0.3cm}
	\label{tab:per_class_avg_modelnet}
\end{table*}

\vspace{-0.1cm}
\subsection{Part Annotation on Real-World Data}
\begin{figure}[t]
	\begin{center}
		\includegraphics[width=1\linewidth]{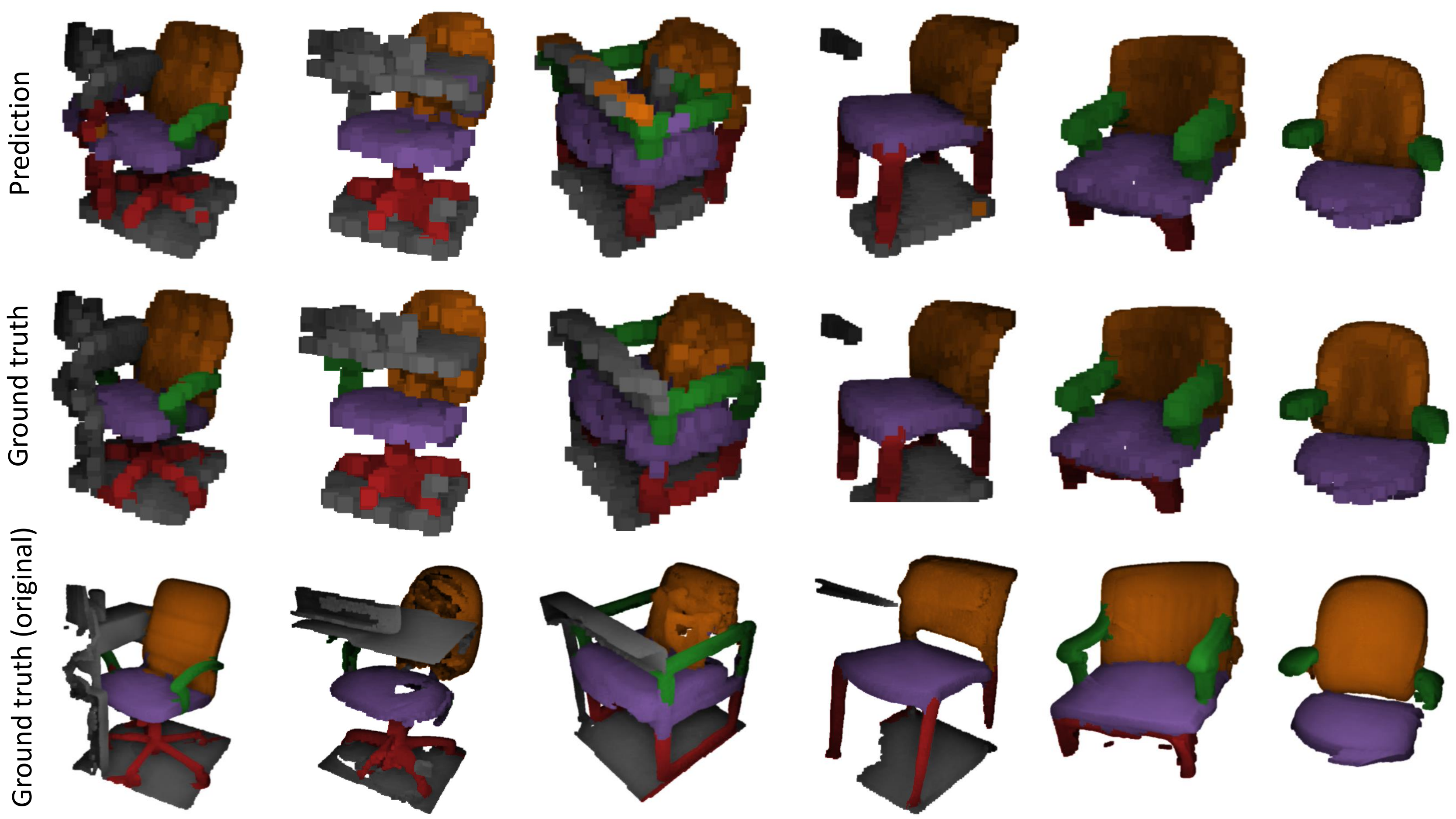}
	\end{center}
	\vspace{-0.2cm}
	\caption{Part segmentation on the chair category. From top to bottom: part prediction, ground truth in 2048 points, and high-resolution ground truth from original point clouds.}
	\vspace{-0.4cm}
	\label{fig:part_seg}
\end{figure}

\vspace{-0.1cm}
We further support part-based annotation in our dataset. So far, point cloud classification methods only evaluate part segmentation task on ShapeNet~\cite{valentin-semanticpaint-tog15}. However, there has been no publicly available dataset for part segmentation on real-world data despite the availability of scene meshes datasets~\cite{hua-scenenn-3dv16,dai2017scannet}. We close this gap with our dataset, which will be released for future research. Figure~\ref{fig:part_seg} shows a visualization of part segmentation on our data. Table~\ref{tab:partseg} and Table~\ref{tab:partseg_perpart} provide a baseline part segmentation evaluation on our data. Using these part annotations may also improve partial object classification in the future. 
%A common approach to partial object classification is to classify based on its parts. So far, point cloud classification methods only evaluate part segmentation task on ShapeNet \cite{valentin-semanticpaint-tog15}. However, there has been no publicly available dataset for part segmentation on real-world data despite the availability of scene meshes datasets~\cite{hua-scenenn-3dv16,dai2017scannet}. We close this gap by supporting part annotation in our dataset. Figure~\ref{fig:part_seg} shows a visualization of part segmentation on our data. Table~\ref{tab:partseg} and Table~\ref{tab:partseg_perpart} provide a baseline part segmentation evaluation on our data. We will release this part annotation for future research upon publication.

\begin{table}[h]
	\begin{center}
		\small
		\begin{tabular}{c|c|c|c|c|c}
			\toprule
			& \rot{OBJ\_BG} & \rot{PB\_T25} & \rot{PB\_T25\_R} & \rot{PB\_T50\_R} & \rot{PB\_T50\_RS}\\
			\midrule
			PointNet \cite{qi-pointnet-cvpr17} & 81.3 & 83.1 & 82.2 & 79.9 & 78.8 \\				
			PointNet++ \cite{qi2017pointnetplusplus} & 80.3 & 85.4 & 84.1 & 81.3 & 82.8 \\
			\bottomrule			
		\end{tabular}
	\end{center}
	\vspace{-0.2cm}
	\caption{Overall accuracy in \% of part segmentation of chairs in the different variants of ScanObjectNN.}
	\vspace{-0.3cm}
	\label{tab:partseg}
\end{table}

\begin{table}[h]
	\begin{center}
		\small
		\begin{tabular}{c|c|c|c|c|c}
			\toprule
			& background & seat & back & base & arm \\
			\midrule
			PointNet \cite{qi-pointnet-cvpr17} & 81.4 & 81.8 & 86.7 & 52.5 & 40.5 \\
			PointNet++ \cite{qi2017pointnetplusplus} & 81.9 & 87.7 & 89.2 & 62.3 & 64.6 \\
			\bottomrule			
		\end{tabular}
	\end{center}
	\vspace{-0.2cm}
	\caption{Per part average accuracy in \% of chairs in our hardest variant PB\_T50\_RS.}
	\vspace{-0.4cm}
	\label{tab:partseg_perpart}
\end{table}

\vspace{-0.1cm}
\subsection{Discussion}
\vspace{-0.1cm}
Our quantitative evaluations show that performing object classification on real-world data is challenging. The state-of-the-art methods in our benchmark have up to 78.5\% accuracy on our hardest variant (PB\_T50\_RS). The benchmark also helps us recognize the following open problems:\vspace{0.2cm}\\
\noindent \textbf{Background} is expected to provide context information but also introduce noise. It is desirable to have an approach that can distinguish foreground from background to effectively exploit context information in the classification task.\vspace{0.1cm}\\
\noindent \textbf{Object partiality}, caused by low reconstruction quality or inaccurate object proposals, also needs to be addressed. Part segmentation techniques~\cite{qi-pointnet-cvpr17,li-pointcnn-ar18} 
%Multi-view representation~\cite{su15mvcnn, yu-multiview-cvpr18, yavar-spnet-accv18} 
could help to describe partial objects.\vspace{0.1cm}\\ 
%However, determining views for point clouds of partial objects is another challenge. In addition, point clouds may be sparse and thus projection of 3D points onto 2D views may produce incomplete shapes.
\noindent \textbf{Generalization} between CAD models and real-world scans needs more investigations. In general, we found that training on real-world data and testing on CADs can generalize better than the opposite case. It could be explained that real-world data have more variations including background and partiality as discussed above. However, CAD models are still important because real-world scans are seldom complete and noise free. Bridging this domain gap could be an important research direction.\\
\vspace{-0.3cm}

To facilitate future work, in the next sections, we propose ideas and baseline solutions. 
\vspace{-0.2cm} 
%Consequently, designing classification networks that are robust to point clouds in the wild could be a potential research direction for point cloud deep learning.

%A summary of all the results are shown in Table~\ref{tab:ours}. As can be seen, by training with our dataset, the best accuracy is improved to about 80\%, which shows that the point cloud classification problem is still far from being fully solved.\\ 
\section{Background-aware Classification Network}
\label{sec:proposednetwork}
\vspace{-0.1cm}
We propose here a simple deep network to handle the occurrence of background in point clouds obtained from real scans; this is one of the open problems we raised in the previous section. An issue with existing point cloud classification networks is the lack of capability to distinguish between foreground and background points. In other words, existing methods take point clouds as a whole and directly calculate features for classification. This issue stems from the design of these networks and also from the simplicity of available training datasets, \eg, ModelNet40. 

To tackle this issue, our idea is to make the network aware of the presence of background by adding a segmentation-guided branch to the classification network. The segmentation branch predicts an object mask that separates the foreground from the background. 
Note that the mask can be easily obtained from our training data since our objects are originally from scene instance segmentation datasets~\cite{hua-scenenn-3dv16,dai2017scannet}. 
\vspace{-0.5cm}
\subsection{Network Architecture}
\begin{figure}[t]
	\begin{center}
		\includegraphics[width=0.8\linewidth]{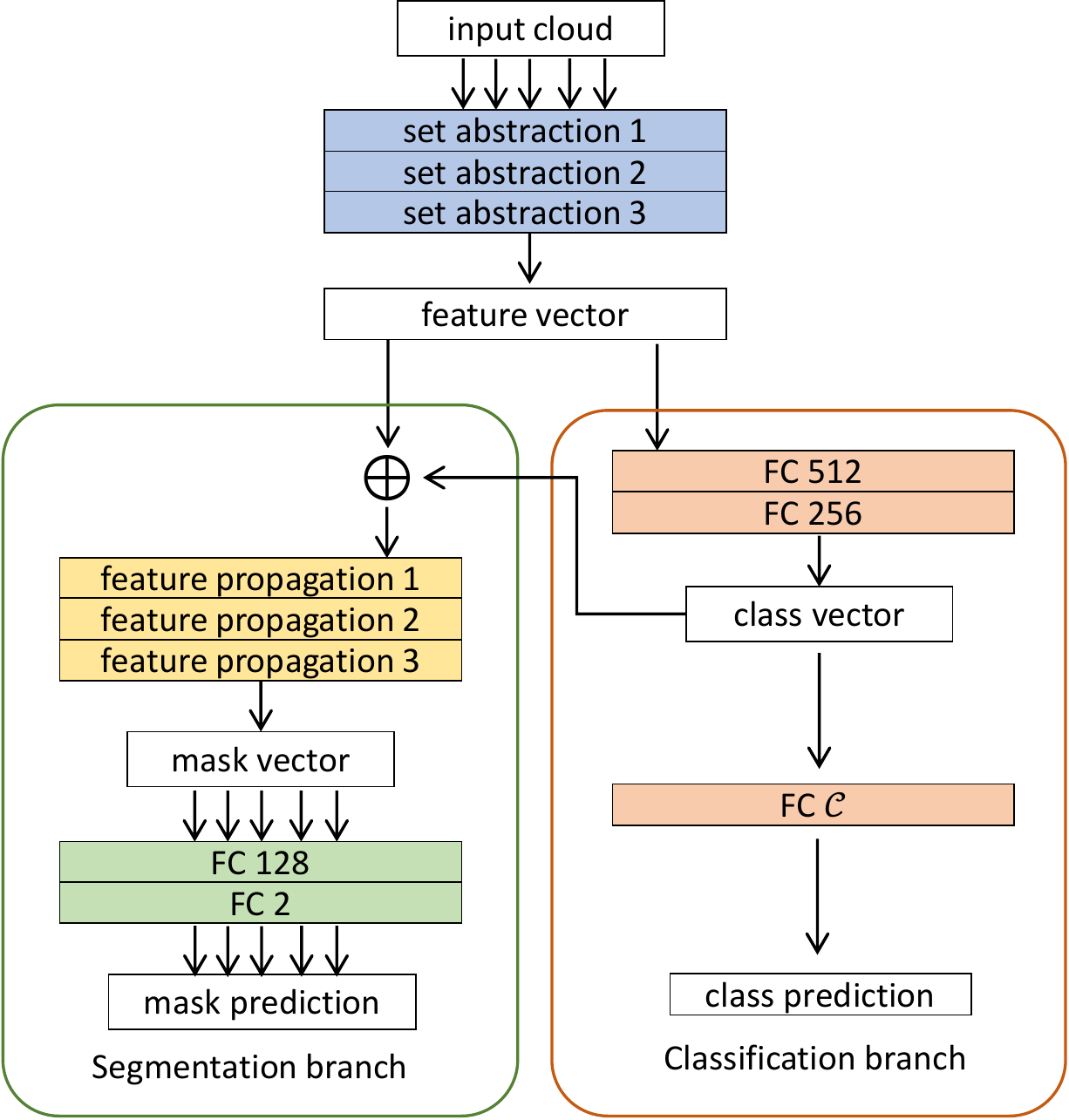}
	\end{center}
	\vspace{-0.2cm}
	\caption{Our proposed network.}
	\vspace{-0.2cm}
	\label{fig:network}
\end{figure}

\vspace{-0.1cm}
Our background-aware (BGA) model is built on top of PointNet++ \cite{qi2017pointnetplusplus} (\textbf{BGA-PN++}). 
Our network is depicted in Figure~\ref{fig:network}.
In particular, we use three levels of set abstractions from the PointNet++ to extract point cloud global features. Global features are then passed through three fully connected layers to produce object classification score. Dropout is also used in a similar manner with the original PointNet++ architecture. Three PointNet feature propagation modules are then employed to compute object masks in segmentation. The feature vector just before the last fully connected layer for the classification score is used as the input to the first PointNet feature propagation modules, making the predicted object mask driven by the classification output. We trained both branches jointly. The loss function is the sum of the classification and segmentation loss, which can be written as 
$\mathcal{L}_{\mathrm{total}} = \mathcal{L}_{\mathrm{class}} + \lambda \mathcal{L}_{\mathrm{seg}}$ where $\mathcal{L}_{\mathrm{class}}$ and $\mathcal{L}_{\mathrm{seg}}$ are both cross entropy losses between the predicted and ground-truth class labels and object masks, respectively. We set $\lambda=0.5$ in our experiments. 

Joint learning for both classification and segmentation with the use of object masks allows the network to be aware of relevant points (\ie, acknowledge the presence of background points). In addition, using classification prediction as a prior to segmentation guides the network to learn object masks that are consistent with the true shape of desired object classes. As to be detailed in our experiments, jointly learning classification and mask prediction results in better classification accuracy in noisy scenarios.

Furthermore, we also introduce \textbf{BGA-DGCNN}, which is a background-aware network based on DGCNN~\cite{wang2018edgeconv}. We apply the same concept as BGA-PN++ that jointly predicts both classification and segmentation, where the last fully connected layer of the classification branch is used as input to the segmentation branch. Our experimental results show that our bga model is adaptive to different network architectures. 

%$\mathcal{L}_{\mathrm{total}} = \lambda \mathcal{L}_{\mathrm{class}} + (1-\lambda) \mathcal{L}_{\mathrm{seg}}$ where $\mathcal{L}_{\mathrm{class}}$ and $\mathcal{L}_{\mathrm{seg}}$ are both cross entropy losses between the predicted and ground-truth class labels and object masks, respectively. We set $\lambda=0.5$ in our experiments. Our network is depicted in Figure~\ref{fig:network}.

\begin{figure}[t]
	\begin{center}
		\includegraphics[width=1\linewidth]{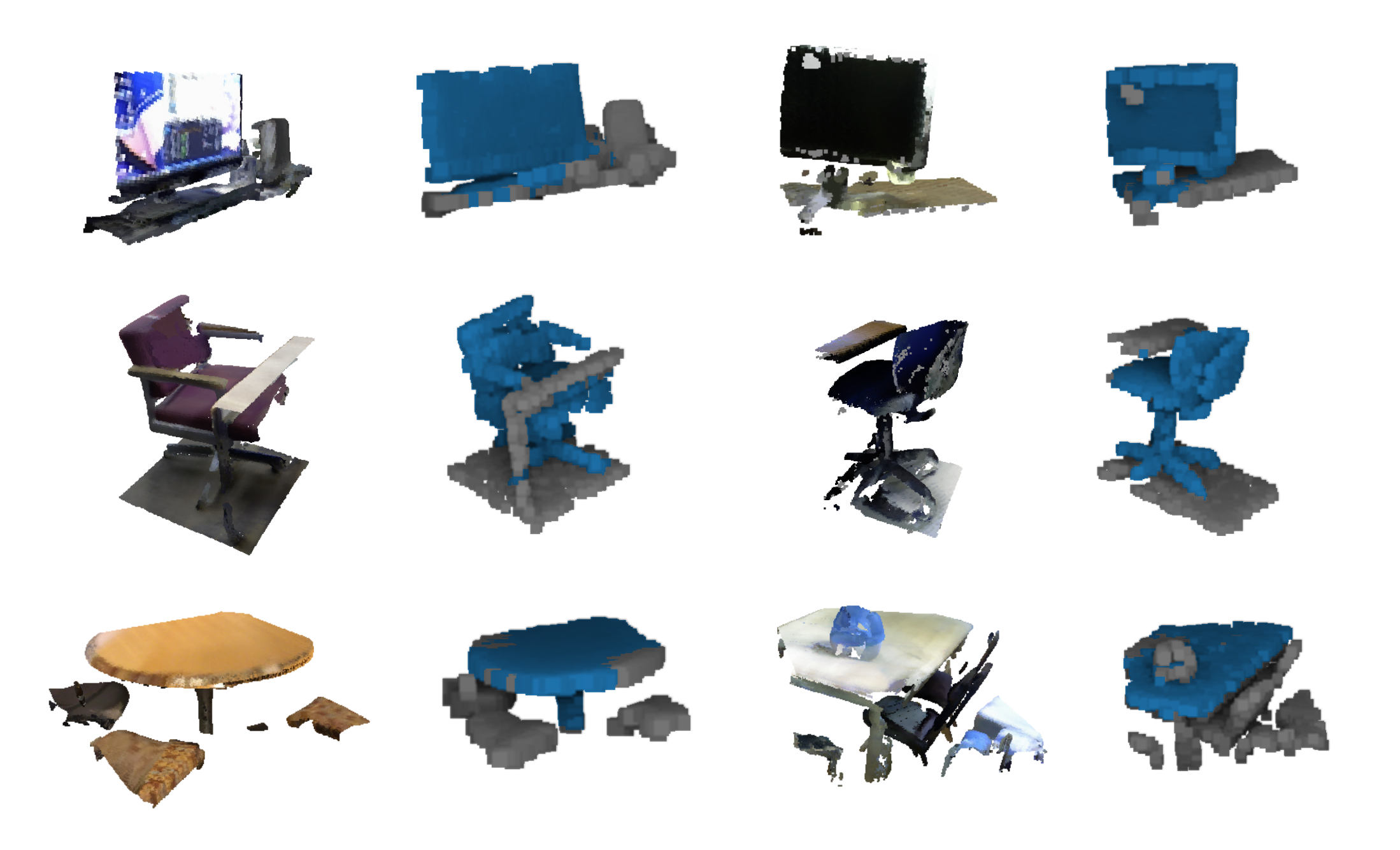}
	\end{center}
	\vspace{-0.3cm}
	\caption{Sample objects and their corresponding predicted masks from the test set of PB\_T50\_RS by our BGA-PN++. Note that color on point clouds is for visualization purposes, but the input to the networks are $(x, y, z)$ coordinates only. }
	\label{fig:mask}
	\vspace{-0.5cm}
\end{figure}

\vspace{-0.1cm}
\subsection{Evaluation}
\vspace{-0.1cm}
We evaluate our network on both our dataset and ModelNet40. Table~\ref{tab:acc_joint} shows a comparison between our network and existing ones on our hardest variant PB\_T50\_RS and ModelNet40 respectively. Our BGA models, BGA-PN++ and BGA-DGCNN, both outperform their vanilla counterparts with BGA-PN++ achieving the best performance on our PB\_T50\_RS. On ModelNet40, our BGA-PN++ improves upon PointNet++ by almost 5\% (with 52.6\% of accuracy), while our BGA-DGCNN achieves the top performance of 56.5\%. Note that, in this evaluation all methods were trained on our \ie PB\_T50\_RS. As shown,
our BGA models gains improvements in both ModelNet40 and our dataset. 

%We evaluate our network on both our dataset and ModelNet40. Table~\ref{tab:acc_joint} shows a comparison between our network and existing ones on our hardest variant PB\_T50\_RS and ModelNet40 respectively. As can be seen, our network with background-awareness capability outperforms existing point cloud-based object classification methods on the hardest variant PB\_T50\_RS of our dataset. On ModelNet40, our network improves upon PointNet++ by almost 5\% (with 52.6\% of accuracy) and is slightly outperformed by DGCNN \cite{wang2018edgeconv} (with 54.7\% of accuracy). Note that, in this evaluation all methods were trained on the same variant, \ie PB\_T50\_RS.

In addition, we also evaluated the segmentation performance of our network. Experimental results showed that our BGA-PN++ performed at 77.6\% and 71.0\%, while our BGA-DGCNN achieved 78.5\% and 74.3\% of segmentation accuracy on our PB\_T50\_RS and ModelNet40, respectively.
We visualize some of the object masks predicted by our BGA-PN++ in Figure~\ref{fig:mask}. It can be seen that our proposed network is able to mask out the background fairly accurately. 
\vspace{-0.2cm}
\subsection{Discussion and Limitation}
\vspace{-0.1cm}
While both BGA models demonstrate good performance, we found that DGCNN-based networks generalizes well between real and CAD data, e.g., when being trained on real and tested on CAD data (Table 9) and vice versa (Table 3). Moreover, Table~\ref{tab:modelnet_to_scan} also show that the same is true for DGCNN-based models on the synthetic to real case. More investigations on the DGCNN architecture could lead to models that generalize better and bridge the gap between synthetic and real data. 

\begin{figure}[t]
	\begin{center}
		\includegraphics[width=\linewidth]{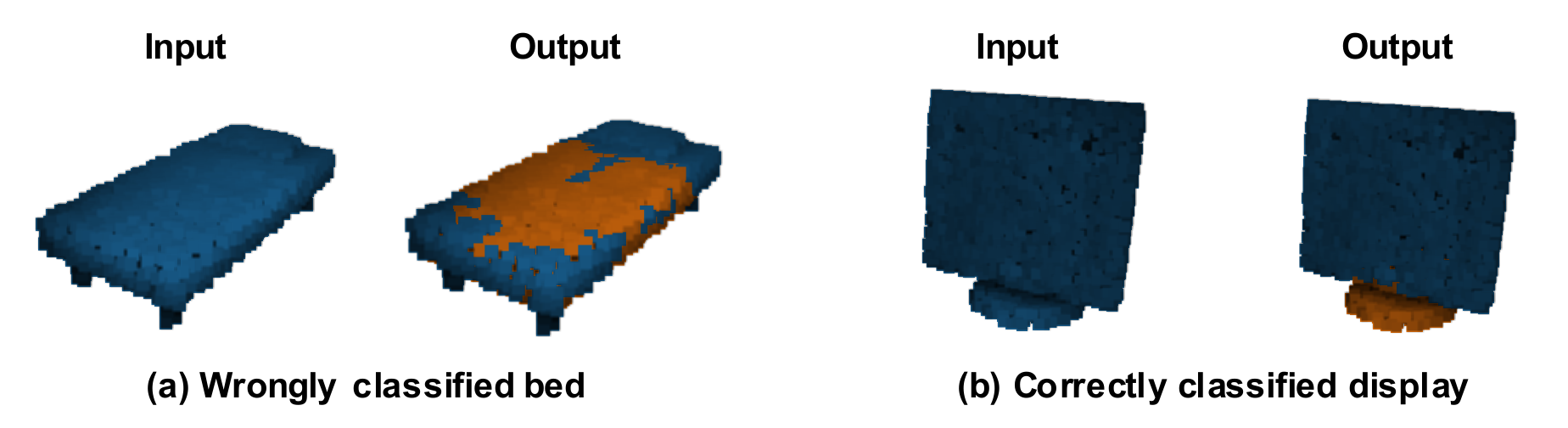}
	\end{center}
	\vspace{-0.4cm}
	\caption{Sample segmentation results of our BGA-PN++ on ModelNet40. Background and foreground are marked in orange and blue, respectively.}
	\label{fig:vis-1}
	\vspace{-0.8cm}
\end{figure}

%Additionally, we also present a limitation of our proposed BGA approach, which is it requires object masks and background points to be included in the data. 
Our proposed BGA is not without limitation. In general, it requires object masks and background to be included in the data. Fig.~\ref{fig:vis-1}-(a) shows a fail case of our method when evaluating on a background-free ModelNet40 object.

\begin{table}
	\begin{center}
		\begin{tabular}{c|c|c|c|c}
			\toprule
			& \multicolumn{2}{c}{Ours} & \multicolumn{2}{c}{ModelNet40} \\
			& OA & mAcc & OA & mAcc \\
			\midrule
			3DmFV \cite{ben20183dmfv} & 63.0 & 58.1 & 51.5 &  52.2\\
			PointNet \cite{qi-pointnet-cvpr17}& 68.2 & 63.4 & 50.9 & 52.7\\
			SpiderCNN \cite{xu2018spidercnn} & 73.7 & 69.8 & 46.6 & 48.8\\	
			PointNet++ \cite{qi2017pointnetplusplus}& 77.9 & 75.4& 47.4 & 45.9\\
			DGCNN \cite{wang2018edgeconv} & 78.1 & 73.6 & 54.7 & 54.9\\
			PointCNN \cite{li-pointcnn-ar18}& 78.5 & 75.1 & 49.2 & 44.6\\
			%\midrule
			BGA-PN++ (ours) & \textbf{80.2} & \textbf{77.5} & 52.6 & 50.6 \\
			BGA-DGCNN (ours) & 79.7 & 75.7 & \textbf{56.5} & \textbf{57.6}  \\
			\bottomrule			
		\end{tabular}
	\end{center}
	\vspace{-0.2cm}
	\caption{Overall and average class accuracy in \% on our PB\_T50\_RS and on ModelNet40. Training is done on our PB\_T50\_RS.}
	\vspace{-0.2cm}
	\label{tab:acc_joint}	
\end{table}

\vspace{-0.2cm}
\section{Conclusion}
\label{sec:conclusion}
\vspace{-0.2cm}
This paper revisits state-of-the-art object classification methods on point cloud data. We found that existing methods were successful with synthetic data but failed on realistic data. To prove this, we built a new real-world object dataset containing $\sim15,000$ objects in 15 categories. Compared with current datasets, our dataset offers more practical challenges including background occurrence, object partiality, and different deformation variants. We benchmarked existing methods on our new dataset, discussed issues, identified open problems, and suggested possible solutions. We also proposed a new point cloud network to classify objects with background. Experimental results showed the advance of our method on both synthetic and real-world object datasets.
\vspace{-1.0cm} 
\paragraph{Acknowledgment} This research project is partially supported by an internal grant from HKUST (R9429). 
%We also sincerely thank all those who helped in data preparation. 

{\small
\bibliographystyle{ieee_fullname}
\bibliography{ref}

\begin{thebibliography}{10}\itemsep=-1pt

\bibitem{armeni_cvpr16}
Iro Armeni, Ozan Sener, Amir~R. Zamir, Helen Jiang, Ioannis Brilakis, Martin
  Fischer, and Silvio Savarese.
\newblock 3d semantic parsing of large-scale indoor spaces.
\newblock In {\em {CVPR}}, 2016.

\bibitem{ben20183dmfv}
Yizhak Ben-Shabat, Michael Lindenbaum, and Anath Fischer.
\newblock 3dmfv: Three-dimensional point cloud classification in real-time
  using convolutional neural networks.
\newblock {\em IEEE Robotics and Automation Letters}, 2018.

\bibitem{bewley2019sim2real}
Alex Bewley, Jessica Rigley, Yuxuan Liu, Jeffrey Hawke, Richard Shen, Vinh-Dieu
  Lam, and Alex Kendall.
\newblock Learning to drive from simulation without real world labels.
\newblock In {\em International Conference on Robotics and Automation (ICRA)},
  2019.

\bibitem{Bobkov2018NoiseResistantDL}
Dmytro Bobkov, Sili Chen, Ruiqing Jian, Muhammad~Z. Iqbal, and Eckehard
  Steinbach.
\newblock Noise-resistant deep learning for object classification in
  three-dimensional point clouds using a point pair descriptor.
\newblock {\em IEEE Robotics and Automation Letters}, 2018.

\bibitem{calli-ycb-robotics17}
Berk Calli, Arjun Singh, James Bruce, Aaron Walsman, Kurt Konolige, Siddhartha
  Srinivasa, Pieter Abbeel, and Aaron~M Dollar.
\newblock Yale-cmu-berkeley dataset for robotic manipulation research.
\newblock {\em International Journal of Robotics Research}, 2017.

\bibitem{Chandler2016MitigationOE}
Ben Chandler and Ennio Mingolla.
\newblock Mitigation of effects of occlusion on object recognition with deep
  neural networks through low-level image completion.
\newblock In {\em Comp. Int. and Neurosc.}, 2016.

\bibitem{chang-shapenet-2015}
Angel~X. Chang, Thomas Funkhouser, Leonidas Guibas, Pat Hanrahan, Qixing Huang,
  Zimo Li, Silvio Savarese, Manolis Savva, Shuran Song, Hao Su, Jianxiong Xiao,
  Li Yi, and Fisher Yu.
\newblock {ShapeNet: An Information-Rich 3D Model Repository}.
\newblock Technical Report arXiv:1512.03012, Stanford University --- Princeton
  University --- Toyota Technological Institute at Chicago, 2015.

\bibitem{choi-dataset-ar16}
Sungjoon Choi, Qian-Yi Zhou, Stephen Miller, and Vladlen Koltun.
\newblock A large dataset of object scans.
\newblock {\em arXiv:1602.02481}, 2016.

\bibitem{dai2017scannet}
Angela Dai, Angel~X Chang, Manolis Savva, Maciej Halber, Thomas Funkhouser, and
  Matthias Niessner.
\newblock Scannet: Richly-annotated 3d reconstructions of indoor scenes.
\newblock In {\em CVPR}, 2017.

\bibitem{mark-sydney-acra13}
Mark~De Deuge, Alastair Quadros, Calvin Hung, and Bertrand Douillard.
\newblock Unsupervised feature learning for classification of outdoor 3d scans.
\newblock In {\em Australasian Conference on Robotics and Automation}, 2013.

\bibitem{domi-feature-wacv18}
M. {Dominguez}, R. {Dhamdhere}, A. {Petkar}, S. {Jain}, S. {Sah}, and R.
  {Ptucha}.
\newblock General-purpose deep point cloud feature extractor.
\newblock In {\em WACV}, 2018.

\bibitem{garcia2017}
Alberto Garcia-Garcia, Jose Rodri­guez, Sergio Orts, Sergiu Oprea, Francisco
  Gomez-Donoso, and Miguel Cazorla.
\newblock A study of the effect of noise and occlusion on the accuracy of
  convolutional neural networks applied to 3d object recognition.
\newblock {\em Computer Vision and Image Understanding}, 2017.

\bibitem{girshick-rcnn-cvpr14}
Ross Girshick, Jeff Donahue, Trevor Darrell, and Jitendra Malik.
\newblock Rich feature hierarchies for accurate object detection and semantic
  segmentation.
\newblock In {\em CVPR}, 2014.

\bibitem{accv2018/Groh}
Fabian Groh, Patrick Wieschollek, and Hendrik P.~A. Lensch.
\newblock Flex-convolution (million-scale point-cloud learning beyond
  grid-worlds).
\newblock In {\em ACCV}, 2018.

\bibitem{han-3d2seqview-tip19}
Z. {Han}, H. {Lu}, Z. {Liu}, C. {Vong}, Y. {Liua}, M. {Zwicker}, J. {Han}, and
  C.~L.~P. {Chen}.
\newblock 3d2seqviews: Aggregating sequential views for 3d global feature
  learning by cnn with hierarchical attention aggregation.
\newblock {\em IEEE Transactions on Image Processing}, 2019.

\bibitem{han-unsupervised-aaai19}
Zhizhong Han, Mingyang Shang, Yu{-}Shen Liu, and Matthias Zwicker.
\newblock View inter-prediction {GAN:} unsupervised representation learning for
  3d shapes by learning global shape memories to support local view
  predictions.
\newblock In {\em AAAI}, 2018.

\bibitem{han-y2seq2seq-aaai19}
Zhizhong Han, Mingyang Shang, Xiyang Wang, Yu{-}Shen Liu, and Matthias Zwicker.
\newblock Y{\^{}}2seq2seq: Cross-modal representation learning for 3d shape and
  text by joint reconstruction and prediction of view and word sequences.
\newblock In {\em AAAI}, 2019.

\bibitem{hermosilla2018monte}
P. Hermosilla, T. Ritschel, P-P Vazquez, A. Vinacua, and T. Ropinski.
\newblock Monte carlo convolution for learning on non-uniformly sampled point
  clouds.
\newblock {\em ACM Transactions on Graphics (Proceedings of SIGGRAPH Asia)},
  2018.

\bibitem{hua-scenenn-3dv16}
Binh-Son Hua, Quang-Hieu Pham, Duc~Thanh Nguyen, Minh-Khoi Tran, Lap-Fai Yu,
  and Sai-Kit Yeung.
\newblock Scenenn: A scene meshes dataset with annotations.
\newblock In {\em International Conference on 3D Vision (3DV)}, 2016.
\newblock \url{http://www.scenenn.net}.

\bibitem{hua-pointwise-cvpr18}
Binh-Son Hua, Minh-Khoi Tran, and Sai-Kit Yeung.
\newblock Pointwise convolutional neural networks.
\newblock In {\em CVPR}, 2018.

\bibitem{hua-objectnn-shrec17}
Binh-Son Hua, Quang-Trung Truong, Minh-Khoi Tran, Quang-Hieu Pham, Asako
  Kanezaki, Tang Lee, HungYueh Chiang, Winston Hsu, Bo Li, Yijuan Lu, Henry
  Johan, Shoki Tashiro, Masaki Aono, Minh-Triet Tran, Viet-Khoi Pham, Hai-Dang
  Nguyen, Vinh-Tiep Nguyen, Quang-Thang Tran, Thuyen~V. Phan, Bao Truong,
  Minh~N. Do, Anh-Duc Duong, Lap-Fai Yu, Duc~Thanh Nguyen, and Sai-Kit Yeung.
\newblock {RGB-D to CAD Retrieval with ObjectNN Dataset}.
\newblock In {\em Eurographics Workshop on 3D Object Retrieval}, 2017.

\bibitem{kanezaki-rotationnet-cvpr18}
Asako Kanezaki, Yasuyuki Matsushita, and Yoshifumi Nishida.
\newblock Rotationnet: Joint object categorization and pose estimation using
  multiviews from unsupervised viewpoints.
\newblock In {\em CVPR}, 2018.

\bibitem{KlokovL17}
Roman Klokov and Victor~S. Lempitsky.
\newblock Escape from cells: Deep kd-networks for the recognition of 3d point
  cloud models.
\newblock 2017.

\bibitem{li-sonet-cvpr18}
Jiaxin Li, Ben~M Chen, and Gim~Hee Lee.
\newblock So-net: Self-organizing network for point cloud analysis.
\newblock In {\em CVPR}, 2018.

\bibitem{li-pointcnn-ar18}
Yangyan Li, Rui Bu, Mingchao Sun, and Baoquan Chen.
\newblock Pointcnn: Convolution on x-transformed points.
\newblock {\em Advances in Neural Information Processing Systems}, 2018.

\bibitem{liu2018point2seq}
Xinhai Liu, Zhizhong Han, Yu-Shen Liu, and Matthias Zwicker.
\newblock Point2sequence: Learning the shape representation of 3d point clouds
  with an attention-based sequence to sequence network.
\newblock {\em arXiv:1811.02565}, 2018.

\bibitem{maturana-voxnet-iros15}
D. Maturana and S. Scherer.
\newblock {VoxNet: A 3D Convolutional Neural Network for Real-Time Object
  Recognition}.
\newblock In {\em {IROS}}, 2015.

\bibitem{nguyen-anno-tvcg17}
Duc~Thanh Nguyen, Binh-Son Hua, Lap-Fai Yu, and Sai-Kit Yeung.
\newblock A robust 3d-2d interactive tool for scene segmentation and
  annotation.
\newblock {\em IEEE Transactions on Visualization and Computer Graphics
  (TVCG)}, 2017.

\bibitem{pham-shrec18}
Quang-Hieu Pham, Minh-Khoi Tran, Wenhui Li, Shu Xiang, Heyu Zhou, Weizhi Nie,
  Anan Liu, Yuting Su, Minh-Triet Tran, Ngoc-Minh Bui, Trong-Le Do, Tu~V. Ninh,
  Tu-Khiem Le, Anh-Vu Dao, Vinh-Tiep Nguyen, Minh~N. Do, Anh-Duc Duong,
  Binh-Son Hua, Lap-Fai Yu, Duc~Thanh Nguyen, and Sai-Kit Yeung.
\newblock {RGB-D Object-to-CAD Retrieval}.
\newblock In {\em Eurographics Workshop on 3D Object Retrieval}, 2018.

\bibitem{qi-pointnet-cvpr17}
Charles~R Qi, Hao Su, Kaichun Mo, and Leonidas~J Guibas.
\newblock Pointnet: Deep learning on point sets for 3d classification and
  segmentation.
\newblock {\em {CVPR}}, 2017.

\bibitem{qi-volumetric-cvpr16}
Charles~R. Qi, Hao Su, Matthias Niessner, Angela Dai, Mengyuan Yan, and
  Leonidas~J. Guibas.
\newblock Volumetric and multi-view cnns for object classification on 3d data.
\newblock In {\em {CVPR}}, 2016.

\bibitem{qi2017pointnetplusplus}
Charles~R Qi, Li Yi, Hao Su, and Leonidas~J Guibas.
\newblock Pointnet++: Deep hierarchical feature learning on point sets in a
  metric space.
\newblock {\em Advances in Neural Information Processing Systems}, 2017.

\bibitem{shen2018kcnet}
Yiru Shen, Chen Feng, Yaoqing Yang, and Dong Tian.
\newblock Mining point cloud local structures by kernel correlation and graph
  pooling.
\newblock In {\em CVPR}, 2018.

\bibitem{nathan-nyu-eccv12}
Nathan Silberman, Derek Hoiem, Pushmeet Kohli, and Rob Fergus.
\newblock Indoor segmentation and support inference from rgbd images.
\newblock In {\em {ECCV}}, 2012.

\bibitem{Simonovsky2017ecc}
Martin Simonovsky and Nikos Komodakis.
\newblock Dynamic edge-conditioned filters in convolutional neural networks on
  graphs.
\newblock In {\em CVPR}, 2017.

\bibitem{singh-bigbird-icra14}
A. {Singh}, J. {Sha}, K.~S. {Narayan}, T. {Achim}, and P. {Abbeel}.
\newblock Bigbird: A large-scale 3d database of object instances.
\newblock In {\em International Conference on Robotics and Automation (ICRA)},
  2014.

\bibitem{song-sunrgbd-cvpr15}
Shuran Song, Samuel~P. Lichtenberg, and Jianxiong Xiao.
\newblock Sun rgb-d: A rgb-d scene understanding benchmark suite.
\newblock In {\em {CVPR}}, 2015.

\bibitem{song-deep-cvpr16}
Shuran Song and Jianxiong Xiao.
\newblock {D}eep {S}liding {S}hapes for amodal 3{D} object detection in {RGB-D}
  images.
\newblock In {\em {CVPR}}, 2016.

\bibitem{su15mvcnn}
Hang Su, Subhransu Maji, Evangelos Kalogerakis, and Erik~G. Learned{-}Miller.
\newblock Multi-view convolutional neural networks for 3d shape recognition.
\newblock In {\em {ICCV}}, 2015.

\bibitem{valentin-semanticpaint-tog15}
Julien Valentin, Vibhav Vineet, Ming-Ming Cheng, David Kim, Jamie Shotton,
  Pushmeet Kohli, Matthias Nie{\ss}ner, Antonio Criminisi, Shahram Izadi, and
  Philip Torr.
\newblock Semanticpaint: Interactive 3d labeling and learning at your
  fingertips.
\newblock {\em {ACM Transactions on Graphics}}, 2015.

\bibitem{wang2018local}
Chu Wang, Babak Samari, and Kaleem Siddiqi.
\newblock Local spectral graph convolution for point set feature learning.
\newblock {\em ECCV}, 2018.

\bibitem{wang2018edgeconv}
Yue Wang, Yongbin Sun, Ziwei Liu, Sanjay~E. Sarma, Michael~M. Bronstein, and
  Justin~M. Solomon.
\newblock Dynamic graph cnn for learning on point clouds.
\newblock {\em arXiv preprint arXiv:1801.07829}, 2018.

\bibitem{wu-3dshapenets-cvpr15}
Zhirong Wu, Shuran Song, Aditya Khosla, Xiaoou Tang, and Jianxiong Xiao.
\newblock 3d shapenets: A deep representation for volumetric shapes.
\newblock In {\em {CVPR}}, 2015.

\bibitem{xu2018spidercnn}
Yifan Xu, Tianqi Fan, Mingye Xu, Long Zeng, and Yu Qiao.
\newblock Spidercnn: Deep learning on point sets with parameterized
  convolutional filters.
\newblock In {\em ECCV}, 2018.

\bibitem{yang-foldingnet-cvpr18}
Yaoqing Yang, Chen Feng, Yiru Shen, and Dong Tian.
\newblock Foldingnet: Point cloud auto-encoder via deep grid deformation.
\newblock In {\em CVPR}, 2018.

\bibitem{yavar-spnet-accv18}
Mohsen Yavartanoo and Euyoung Kim.
\newblock Spnet: Deep 3d object classification and retrieval using
  stereographic projection.
\newblock In {\em ACCV}, 2018.

\bibitem{you-joint-mm18}
Haoxuan You, Yifan Feng, Rongrong Ji, and Yue Gao.
\newblock Pvnet: A joint convolutional network of point cloud and multi-view
  for 3d shape recognition.
\newblock In {\em Proceedings of the ACM International Conference on
  Multimedia}, 2018.

\bibitem{yu-multiview-cvpr18}
Tan Yu, Jingjing Meng, and Junsong Yuan.
\newblock Multi-view harmonized bilinear network for 3d object recognition.
\newblock In {\em CVPR}, 2018.

\bibitem{zaheer-deepsets-nips17}
Manzil Zaheer, Satwik Kottur, Siamak Ravanbakhsh, Barnabas Poczos, Ruslan~R
  Salakhutdinov, and Alexander~J Smola.
\newblock Deep sets.
\newblock In {\em Advances in Neural Information Processing Systems}, 2017.

\end{thebibliography}
}

\end{document}